\PassOptionsToPackage{dvipsnames,table,xcdraw}{xcolor}
\documentclass[10pt,twocolumn,letterpaper]{article}

\usepackage{cvpr}              
\usepackage[table]{xcolor}

\newcommand{\modelname}{GSGTrack}
\usepackage{multirow}
\usepackage{makecell}
\usepackage[normalem]{ulem}
\usepackage{setspace}
\usepackage{pifont}
\usepackage{float}
\usepackage{color}
\usepackage[dvipsnames]{xcolor}

\definecolor{cvprblue}{rgb}{0.21,0.49,0.74}
\usepackage[pagebackref,breaklinks,colorlinks,allcolors=cvprblue]{hyperref}

\def\paperID{4953} 
\def\confName{CVPR}
\def\confYear{2025}

\title{GSGTrack: Gaussian Splatting-Guided Object Pose Tracking from RGB Videos}

\author{Zhiyuan Chen\textsuperscript{1}, Fan Lu\textsuperscript{1}, Guo Yu\textsuperscript{1}, Bin Li\textsuperscript{1}, Sanqing Qu\textsuperscript{1},\\ Yuan Huang\textsuperscript{2}, Changhong Fu\textsuperscript{1}, Guang Chen\textsuperscript{1}\thanks{Corresponding author: guangchen@tongji.edu.cn}\\
{\small $^{1}$Tongji University, $^{2}$Control science and engineering, Beijing Institute of Control Engineering}}
\begin{document}
\maketitle
\begin{abstract}


Tracking the 6DoF pose of unknown objects in monocular RGB video sequences is crucial for robotic manipulation. However, existing approaches typically rely on accurate depth information, which is  non-trivial to obtain in real-world scenarios. Although depth estimation algorithms can be employed,  geometric inaccuracy can lead to failures in RGBD-based pose tracking methods. To address this challenge, we introduce \modelname, a novel RGB-based pose tracking framework that jointly optimizes geometry and pose. Specifically, we adopt 3D Gaussian Splatting to create an optimizable 3D representation, which is learned simultaneously with a graph-based geometry optimization to capture the object's appearance features and refine its geometry.  However, the joint optimization process is susceptible to perturbations from noisy pose and geometry data. Thus, we propose an object silhouette loss to address the issue of pixel-wise loss being overly sensitive to pose noise during tracking. To mitigate the geometric ambiguities caused by inaccurate depth information, we propose a geometry-consistent image pair selection strategy, which filters out low-confidence pairs and ensures robust geometric optimization. Extensive experiments on the OnePose and HO3D datasets demonstrate the effectiveness of \modelname 
 in both 6DoF pose tracking and object reconstruction.











%
%

\end{abstract}    
\section{Introduction}
\label{sec:intro}








6DoF object pose tracking aims to continuously estimate the 3D position and orientation of target objects from consecutive video sequences. This provides consistent and accurate positional information for objects being manipulated, which is essential for applications such as robotic manipulation~\cite{kappler2018real,wen2022catgrasp} and control planning~\cite{Sleiman2021UnifiedMPC}.


Early 6DoF object pose estimation or tracking approaches assume access to 3D models~\cite{Xiang2018PoseCNN,Peng2019PVNet} or category templates~\cite{Wang2019NOCS,Hinterstoisser2012ModelBased} and rely on feature matching algorithms to estimate the pose of the target object~\cite{Pitteri2019CorNet}. This reliance makes it extremely challenging for the models to generalize to novel, unseen objects.
To achieve pose tracking of unknown objects, some studies have extended the concept of online localization from SLAM algorithms to pose tracking tasks~\cite{Yang2023UniQuadric,Wen2023BundleSDF}. Given an RGBD video sequence, they project the 2D object into a 3D point cloud using accurate depth information, and employ point cloud registration to track the 6DoF pose. This pipeline inherently relies on accurate depth information for pose estimation. However, on most lightweight robots equipped with monocular vision systems, obtaining accurate depth information is usually not feasible, which poses significant challenges for the application of pose tracking algorithms~\cite{Ganj2023MobileARDepth}.

\begin{figure}
    \centering
    \includegraphics[width=0.95\linewidth]{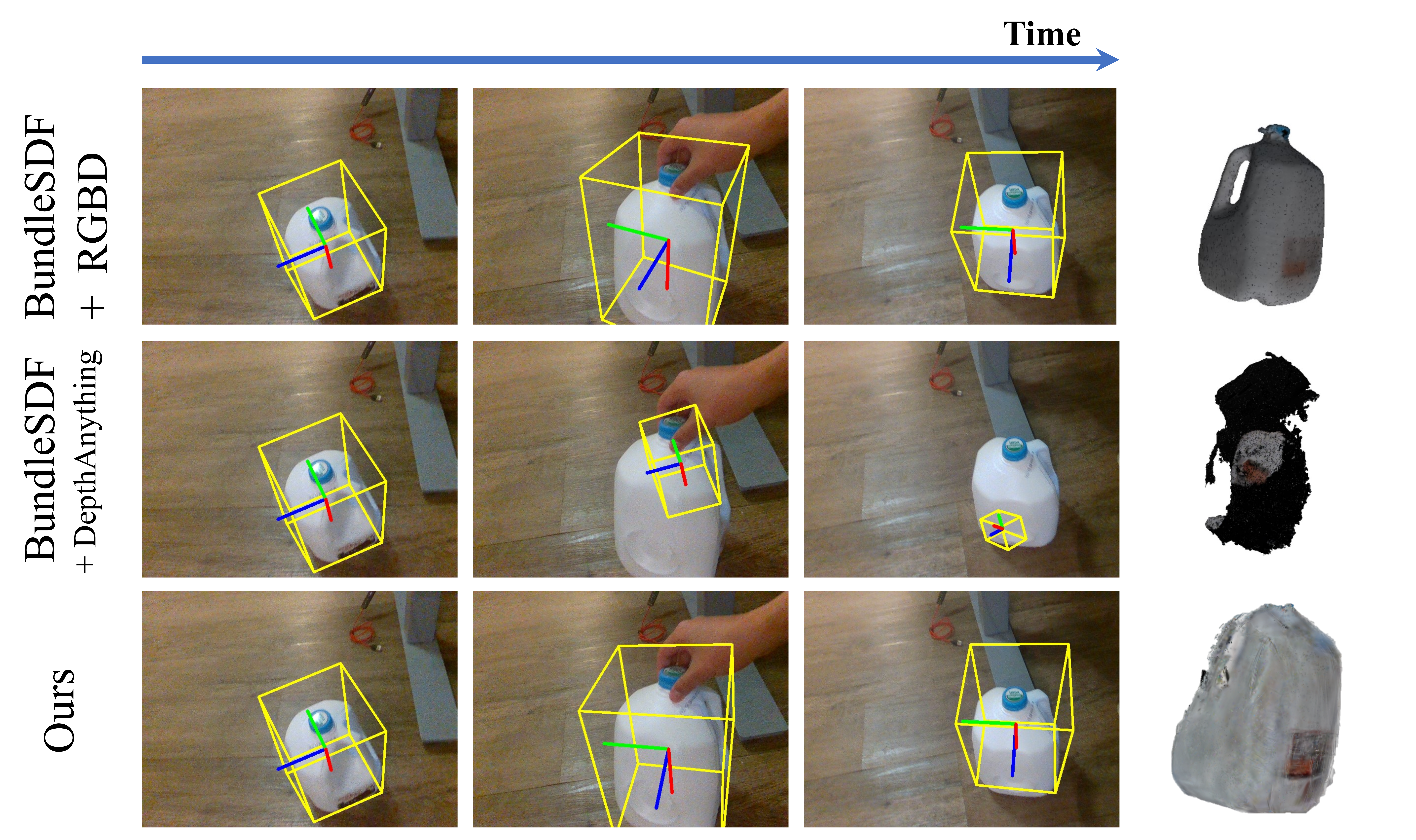}
    \caption{We are tackling a challenging problem: tracking 6DoF pose of unknown objects from RGB videos without accurate depth information. When applied to RGB videos with inaccurate estimated depth information~\cite{Yang2024DepthAnything}, RGBD-based methods~\cite{Wen2023BundleSDF} degenerates quickly. In contrast, our method achieves robust tracking and reconstruction results.}
    \label{fig:Teaser}
    \vspace{-4ex}
\end{figure}

To achieve 6DoF pose tracking using monocular RGB videos without depth information, a straightforward alternative is to use monocular depth estimation methods to obtain depth information~\cite{Zhao2023LiteMono,Yang2024DepthAnything}.
However, previous RGBD-based methods are fragile to depth noise~\cite{Wen2023BundleSDF,wen2021bundletrack,Yang2023UniQuadric}.  As shown in \cref{fig:Teaser}, noisy depth information can lead to increased coarse pose errors in registration-based methods and introduce incorrect target information during pose optimization, resulting in a quick degeneration.

To this end, we propose a Gaussian Splatting Guided object pose tracking framework, termed GSGTrack, which achieves RGB-based 6DoF object pose tracking by jointly optimizing pose and geometry. Specifically, we represent the object using 3D Gaussian Splatting (3DGS)~\cite{kerbl2023gaussian}, reformulating it as an online reconstruction pipeline that continuously reconstructs the object while guiding pose optimization through rendering losses. To enable accurate and robust object pose tracking even with inaccurate initial geometry, we design a graph-based geometric optimization method that jointly optimizes both poses and the 3D representation via an online geometric structure graph. However, the optimization process remains prone to noise in poses and geometry. To mitigate this, we design a differential silhouette loss and an outlier image pair pruning strategy, which leverages confidence metrics from pixel depth predictions, pose deviation from inertial data, and similarity between new and historical image geometry, enabling pruning of mismatched image pairs based on 3D geometric consistency.




To evaluate the proposed method, extensive experiments are conducted on two monocular RGB object pose tracking datasets, \ie, OnePose dataset~\cite{lin2022onepose}  and HO3D dataset~\cite{hampali2020honnotate}. The results demonstrate that the proposed method significantly outperforms existing approaches in terms of both accuracy and reconstruction quality. To summarize, our main contributions are as follows:





\begin{itemize}
    \item  We propose a novel framework for monocular RGB-basd 6DoF object pose tracking, which operates online and achieves robust pose tracking even with inaccurate geometric structures.

    \item  We introduce a confidence-based pruning and optimization method for image pairs, effectively mitigating the impact of abnormal registration results on the global model. 

    \item  Extensive experiments show that GSGTrack significantly outperforms state-of-the-art methods on monocular RGB video object pose tracking datasets.
\end{itemize}

\section{Related Work}
\label{sec:related}


\noindent \textbf{6-DoF Object Pose Estimation and Tracking.} Estimating the 6DoF pose of an object directly from RGB images is inherently an ill-posed problem, often requiring additional 3D information to resolve ambiguities. One approach introduces CAD models for offline training~\cite{kehl2017sd6d,labbe2020cosypose,park2019pix2pose}, but this limits generalization to novel objects. Some methods attempt to relax this with category templates~\cite{Wang2019NOCS,Hinterstoisser2012ModelBased}, yet their performance depends heavily on template accuracy, posing practical challenges.
Other approaches leverages new view synthesis methods, such as Mask-RCNN~\cite{cai2020reconstruct}, NeRF~\cite{li2022nerfpose}, or 3DGS~\cite{cai2024gspose,luo2024sgpose}, to incorporate 3D shape information, achieving CAD-free pose tracking. However, they still require pre-captured reference views of the test object, which can be impractical in many scenarios.
In 6DoF pose tracking, temporal information is used to estimate object poses across video frames. Some studies propose constructing 3D models from multi-view video frames to extend tracking to unknown objects~\cite{Wen2023BundleSDF,wen2021bundletrack}. BundleSDF~\cite{Wen2023BundleSDF} is most similar to our approach, achieving pose tracking and reconstruction for unseen objects. Our method, however, integrates tracking and reconstruction with shape acquisition and optimization, enabling both accurate object pose tracking in RGB videos and improved appearance reconstruction. We further enhance reconstruction quality by incorporating generalized stereo matching~\cite{wang2024dust3r} as a 3D prior.


\begin{figure*}[ht]
\centering
  \includegraphics[width=1\textwidth]{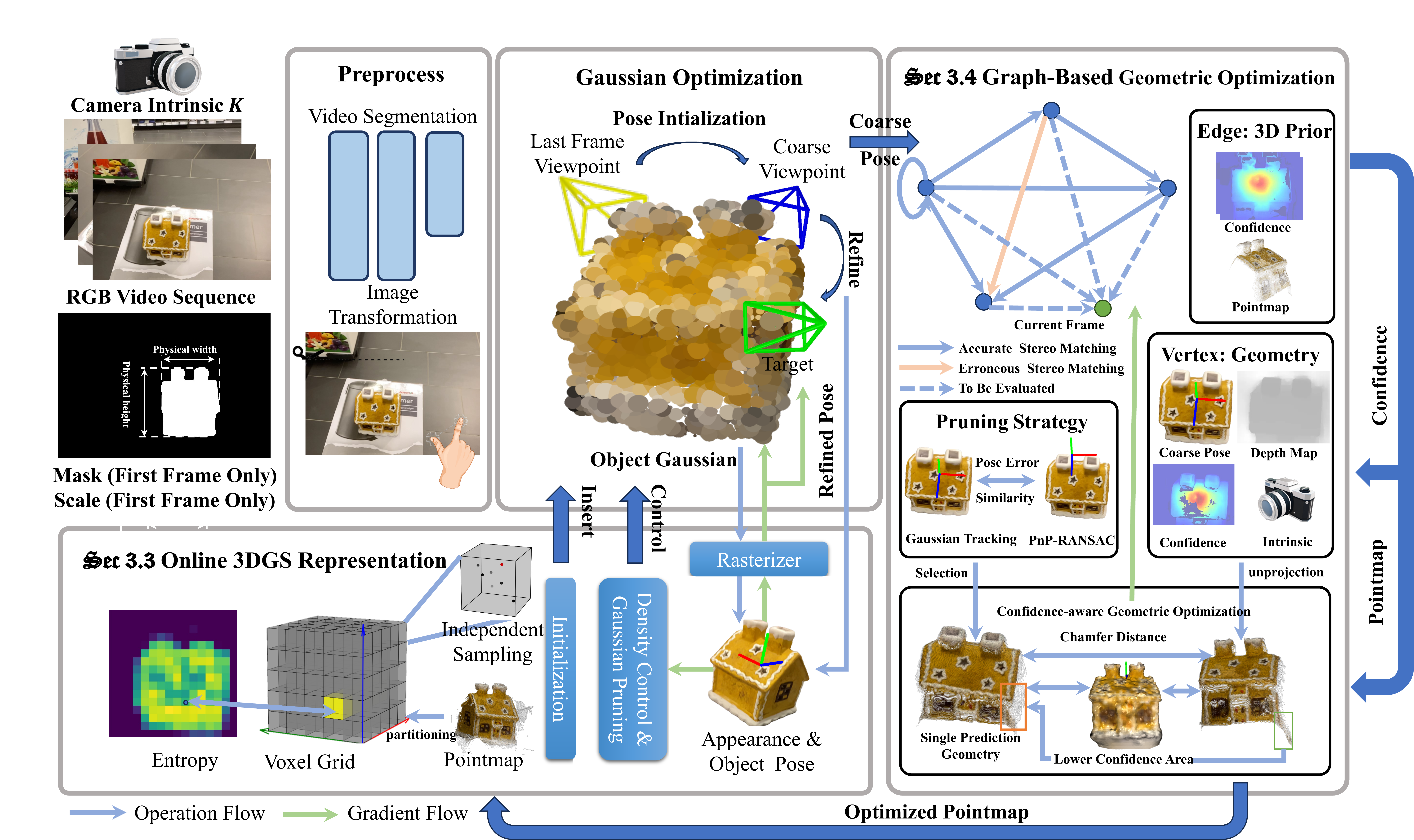}
  \caption{\textbf{Overview of our proposed GSGTrack.} To achieve accurate 6DoF object pose tracking without relying on precise depth information, we propose a joint optimization framework. Starting with a video sequence, we preprocess consecutive frames by generating object masks and estimating coarse geometry. Next, we introduce an online 3DGS representation that facilitates continuous object reconstruction from incoming video frames. Building on this 3D representation, we design a graph-based geometric optimization framework that refines both object pose and 3D structure through an online geometric structure graph. Additionally, we introduce an image pair pruning strategy and a confidence-aware geometric optimization technique to enhance the robustness and accuracy of the optimization process.}
  \label{fig:overview}
\end{figure*}  

\vspace{0pt}\noindent \textbf{Simultaneous Localization and Mapping Algorithms.} RGB-SLAM algorithms primarily achieve camera pose estimation and scene map reconstruction from monocular RGB video sequences, addressing a problem similar to ours~\cite{mur2017orb,klein2007parallel}. However, RGB-SLAM algorithms are mainly applied to localization and mapping in large static scenes. Although some variants have extended SLAM to dynamic scenes~\cite{bescos2018dynaslam,bescos2020dynaslam2}, these approaches typically mask dynamic objects from the scene, using the static portions to estimate camera poses and reconstruct the scene map. This limitation prevents them from handling the reconstruction of dynamic objects within a scene.Additionally, other research has introduced the concept of Object SLAM, where algorithms not only reconstruct the scene representation but also detect and recognize the semantics and basic appearance of objects~\cite{yang2019cubeslam,nicholson2018quadricslam}. However, these algorithms cannot address scenarios involving dynamic interactions between objects and the environment, nor can they fully achieve 3D reconstruction of objects. In contrast, our method utilizes an innovative 3D Gaussian Splatting based representation technique . By integrating newly observed RGB images with existing Gaussian spheres, our approach generates a consistent 3D representation while simultaneously recovering 6DoF pose information of the object.


\vspace{0pt}\noindent \textbf{3D Reconstruction.} Reconstructing 3D representations from 2D images has been widely studied with learning-based methods~\cite{choy20163d,wang2018pixel2mesh,yao2018deepmvs}. Recent advances in neural scene representations now allow high-quality 3D reconstructions~\cite{mildenhall2020nerf,kerbl2023gaussian}, though they typically assume known camera poses, limiting applicability. Some pose-free 3D reconstruction methods have emerged, but they focus mainly on static scenes, making them unsuitable for dynamic object interactions~\cite{fu2024colmapfree,lin2021barf,ye2024noposplat}. In particular, BundleSDF proposes an online approach that reconstructs 3D object meshes using SDF, relying on accurate depth information from RGBD images, but with limited appearance reconstruction~\cite{Wen2023BundleSDF}. In contrast, our method uses 3D Gaussian Splatting, supervised by RGB images, to achieve stronger appearance reconstruction. Another line of research leverages end-to-end feedforward neural networks for 3D scene reconstruction~\cite{zhang2021neuralrecon}. For example, Dust3R and similar methods~\cite{wang2024dust3r,fan2024instantsplat,weinzaepfel2024splattr} use generalized stereo networks to generate point clouds, while some 3D Gaussian Splatting approaches directly predict Gaussian attributes~\cite{zhang2023drivefeedforward}. However, these rely on accurate priors and are less effective in dynamic scenes. Our method instead employs a selective geometric optimization strategy, addressing challenges when priors are imprecise.

\section{Methodology}

\subsection{Preliminary}
\noindent \textbf{Problem Formulation.}  In an object-centric dynamic scene, given a collected monocular RGB video sequence $F = \left\{F_{0}, F_{1}, \ldots, F_{n-1}\right\}\left(F_{t} \in \mathbb{R}^{W \times H \times 3}\right)$, along with the segmentation mask o $M_{0}$ and the ground truth projected size $\mathbf{s}_{0}=\left[\begin{array}{l}w_{0} \\h_{0}\end{array}\right] \in \mathbb{R}^{2}$   of the object \textit{in the first frame only} as inputs, the goal of GSGTrack is to track the 6DoF pose of the object online while reconstructing a textured 3D model of the object.

\noindent \textbf{Perliminary for 3DGS~\cite{kerbl2023gaussian}.} As mentioned before, we use 3D Gaussian Splatting (3DGS) as our basic object representation. 3DGS is a differential 3D representation for real-time neural rendering. Thanks to the explicit representation, 3DGS enables fast 3D scene reconstruction and rendering, making it suitable as a basic 3D representation for object pose tracking. Specifically, 3DGS represents a scene as a set of anisotropic 3D Gaussian sphere. Each gaussian sphere is defined with a center $\mu_p$, a covariance matrix $\Sigma$, a view-dependent color $c$, and a transparency $\alpha$. For rendering, 3DGS project all 3D Gaussian spheres into 2D Gaussian distributions through a differentiable Gaussian splatting pipeline, and then blend the colors using fast alpha blending. The rendering process can be summarized as follows:
\begin{equation}
    \textstyle
    \boldsymbol{\mu}^{\prime}=\pi\left(\boldsymbol{T} \cdot \boldsymbol{\mu}\right), \quad
    \boldsymbol{\Sigma}^{\prime}=J W \boldsymbol{\Sigma} W^{T} J^{T},
    \label{eq:project_2d_gaussian}
\end{equation}
\begin{equation}
    \textstyle
    C=\sum_{i \in M} \mathcal{C}_{i} \alpha_{i} \prod_{j=1}^{i-1}\left(1-\alpha_{i}\right),
    \label{eq:color_rendering}
\end{equation}
where $\pi$ is the projection operation, $T$ is the camera pose of the viewpoint,  $W$ is the rotational part of $T$ and $J$.





\subsection{Joint Optimization Framework}
To achieve accurate 6DoF object pose tracking and reconstruction under noisy geometric information, we propose a joint optimization framework, which jointly optimizes object poses and 3D representation. Specifically, given consecutive video frames, we employ generalized stereo matching network~\cite{wang2024dust3r} to estimate coarse geometric information. The estimated results include dense, pixel-aligned 3D pointmaps $X^{u}_{e}$ and $X^{v}_{e}$ in a shared local coordinate system $O_e$ and also confidence maps $C^{u}_{e}$ and $C^{v}_{e}$. Then, we introduce an Online 3DGS representation to enable continuous, real-time object reconstruction even with noisy poses and imprecise initial points (\cref{sec:object_gaussian}). To track object pose, for each video frame $F_t$, we first compute and optimize its 6DoF pose relative to the 3DGS by performing pose optimization where gradients are propagated solely to the pose parameters. This estimated pose serves as a coarse initialization for subsequent pose refinement and object reconstruction. Each frame is then integrated into an online geometric structure graph, where an image pair pruning strategy and a confidence-aware geometric optimization strategy are employed to fuse geometrically accurate historical frame information to estimate the geometry of the current frame  (\cref{sec:graph}). Then, we simultaneously optimize the 3DGS and refine the object pose by incorporating both photometric and depth losses. Details of the gradient flow is provided in our supplementary material.

\subsection{Online 3DGS Representation}
\label{sec:object_gaussian}

Unlike traditional 3DGS, which reconstructs scenes from a fixed set of images, our approach performs reconstruction as an ongoing process, continuously incorporating new object views. To enable this, we developed an online 3DGS framework that supports dynamic Gaussian insertion and pruning under noisy pose and inaccurate points.


\noindent \textbf{Gaussian Insertion.} Due to the redundancy and high error rate in the initial pointmap~\cite{wang2024dust3r}, it is unsuitable for directly initializing Gaussian spheres. Conventional downsampling methods based on confidence ~\cite{fan2024instantsplat} or random sampling often lead to gaps in the training results. To address this challenge, we propose a downsampling method that incorporates image complexity. Specifically, the 3D space is divided into a $K \times K \times K$ voxel grid, and the corresponding 2D image is partitioned into $K \times K$ squares.  As shown in Eq.~\eqref{eq:entropy}, we compute the image entropy $E$ for each region to determine the number of sampling points $ N \propto E $ for each voxel column.

\begin{small}
\setlength\abovedisplayskip{0.cm}
    \begin{equation}
        E_{ij} = - \sum_{p=0}^{L-1} P_{ij}(p) \cdot \log_2 \left( P_{ij}(p) \right),
        \label{eq:entropy}
    \end{equation}
\end{small}
where $L$ represents the number of grayscale levels, and $P_{ij}(p)$ denotes the probability of a pixel having grayscale value $p$ within the region $\text{block}_{ij}$.

Given that the pointmap is generated through unprojection from a dense 2D depth map, points in each voxel column typically fall within the same voxel. To ensure a rich hierarchical structure in the sampled pointmap and to enhance the 3D representation capability of Gaussian spheres, we perform random pointmap interpolation along the negative direction of each point’s normal vector within the pointmap. The maximum number of sampled points per voxel is set to $K/2$. Within each voxel, points are sampled randomly, with the sampling probability of each point proportional to its confidence level, which can be obtained form the generalized stereo matching network.

\noindent \textbf{Gaussian Optimization.} During the optimization of the 3DGS, we utilize RGB image to guide the photometric optimization of the rendering results. The photometric loss $\mathcal{L}_p$ can be represented as:

\begin{small}
\setlength\abovedisplayskip{0.cm}
\begin{equation}
    \mathcal{L} _{p}=\left\|I\left(\mathcal{G}, \boldsymbol{T}\right)-{I}_{gt}\right\|_{1},
    \label{eq:pho loss}
\end{equation}
\end{small}
where $I\left(\mathcal{G}, \boldsymbol{T}\right)$ represents the rendering result of the Gaussian model $\mathcal{G}$ from the viewpoint $\boldsymbol{T}$, while $I_{gt}$ denotes the ground truth image.

We concurrently use the depth map from geometric optimization, which will be introduced in \cref{sec:graph}, to guide the geometric refinement of the 3D representation. This depth map is generated via alpha blending of depth data from Gaussian spheres. Unlike the depth losses in algorithms like SparseGS~\cite{xiong2023sparsegs} and GS-SLAM~\cite{matsuki2024gaussian}, we address inherent errors in depth supervision signals by incorporating a depth confidence map derived from geometric structure optimization. The depth loss $\mathcal{L}_d$ can be formally represented as:

\begin{small}
\setlength\abovedisplayskip{0.cm}
    \begin{equation}
        \mathcal{L}_{D} = \sum_{p \in \Omega}C_p \cdot (\sum_{i \in \mathcal{N}} z_{i} \alpha_{i} \prod_{j=1}^{i-1}\left(1-\alpha_{j}\right))-D_{gt}^p,
    \end{equation}
\end{small}
where $\Omega$ represents the set of pixels, \( z_i \) represents the depth of the Gaussian sphere \( i\), and \( \alpha \) represents the transparency of the Gaussian sphere \( i\).

\noindent \textbf{Gaussian Pruning.} To address geometric inaccuracies in initial pointmaps, inspired by~\cite{matsuki2024gaussian}, we prune erroneous points by employing a simple mask-based approach. Specifically, after each round of model training on $F_t$, we select several reference frames from past frames and remove any newly added Gaussian points that project outside the mask in the reference frames.

\subsection{Graph-based Geometric Optimization}\label{sec:graph}
Geometric optimization seeks to achieve precise object pose tracking, even with imperfect geometric information. To accomplish this, we first perform object pose tracking and then construct an online geometric structure graph to further refine the tracked pose. In this process, we employ an image pair pruning strategy and confidence-aware geometric optimization to enhance the accuracy of the optimization.

\noindent\textbf{Pose Initialization.} 
A proper initial pose is essential for effective pose optimization and object reconstruction. Previous methods typically rely on point cloud registration to obtain the initial pose; however, this approach often fails due to noise in the initial geometry. To address this, we propose an geometry-based strategy for object pose initialization. Specifically, we first compute the relative pose between the current frame $F_{t}$ and the object's Gaussian representation $\mathcal{G}_{t-1}$, initializing the coarse pose $\tilde{\xi}_{t}$ with the pose $\xi_{t-1}$ from the previous frame.

The process outlined above provides an initial pose estimation; however, this initial pose is often coarse and inaccurate. To refine the pose, we leverage image texture information for pose optimization. A straightforward approach is to minimize photometric loss $\mathcal{L}_p$; however, relying solely on photometric loss often causes pose optimization to converge on viewpoints with similar textures, resulting in significant errors. To address this, we propose incorporating silhouette loss to capture the object's geometric structure more accurately. While a typical approach is to calculate the Intersection over Union (IoU) between visible and segmented silhouettes~\cite{zhang2023shape,di2017reconstruction}, this method can lack gradients when overlap is minimal. To overcome this limitation, we introduce a distance-based metric that weights the loss by computing each pixel's distance to the nearest silhouette edge. Our optimized silhouette loss is therefore formulated as:

\begin{small}
\setlength\abovedisplayskip{0.cm}
\begin{equation}
\mathcal{L}_{\text{s}} = \frac{1}{|\Omega|} \sum_{p \in \Omega} \left( D_S(p) \cdot (1 - \tilde{S}(p)) + D_{\tilde{S}}(p) \cdot (1 - S(p)) \right),
\end{equation}
\end{small}
where $S$ and $\tilde{S}$ denote the binary masks for the ground truth and rendered image, $D_S$ and $D_{\tilde{S}}$ correspond to the Euclidean distance transforms.






\noindent\textbf{Online Geometric Structure Graph.} As we mentioned before, we construct a geometric structure graph $\mathcal{H}$ for pose optimization. $\mathcal{H}$ is a directed graph, where each node $v$ represents the geometric structure of the image frame $F_v$, including the 6DoF pose $T_v$ and the 3D representation; each edge represents the results from the 3D generalized stereo matching network~\cite{wang2024dust3r}, including pixel-aligned 3D pointmaps $X^{u}_{e}, X^{v}_{e}$ and confidence maps $C^{u}_{e}, C^{v}_{e}$ of two frames. The graph $\mathcal{H}$ stores rich historical information to avoid tracking drift, besides, the storage of matching results also avoid repetitive computation.



 \noindent\textbf{Image Pair Pruning Strategy.} To mitigate long-term tracking drift due to catastrophic forgetting, it is essential to store historical frame information and use multiple frames to jointly predict current frame geometry for accurate pose tracking and reconstruction. However, the 3D prior knowledge obtained through generalized stereo matching network predictions may be inaccurate. To address this issue, we propose an image pair pruning strategy, which removes mismatched results that do not satisfy geometric consistency from the global geometric structure graph. Specifically, we design three different strategies as following:


\noindent \textit{(1) Pose consistency-based pruning:}
To mitigate significant pose errors caused by object symmetry, we use the PnP-RANSAC algorithm to estimate the relative pose between the current frame and the reference frame within the local coordinate system. This estimated pose is then compared to the coarse pose obtained from tracking, with image pairs discarded if mismatches exceed the rotation threshold $\tau_{r}$ or the translation threshold $\tau_{t}$.

\noindent \textit{(2) Geometry similarity-based pruning:} To address geometry mispredictions in low-texture regions, we compare the predicted geometry of the reference frame with the actual reference structure at image pair nodes. This comparison uses the Chamfer Distance (CD) between point clouds to evaluate shape similarity and we filter out edges with low similarity to the reference nodes.

\noindent \textit{(3) Pixel credibility-based edge cropping:} 
We calculate the confidence of edge $(u,v)$ based on the confidence maps produced by the generalized stereo matching network:

\begin{small}
\vspace{-2ex}
\setlength\abovedisplayskip{0.cm}
    \begin{equation}
       \mu=\frac{1}{w \cdot h} \left(\sum_{i=1}^{w} \sum_{j=1}^{h}{C_{u}^{(i,j)} \cdot M_{u}^{(i,j)}} \times \sum_{j=1}^{h}{C_{v}^{(i,j)} \cdot M_{v}^{(i,j)}}\right),
       \label{eq:confidence}
    \end{equation}
\end{small}
where $\mu$ is the edge confidence, $C$ is the confidence map, and $M$ is the segmentation mask. A confidence threshold hyperparameter $\tau_{c}$ is introduced to prune edges that were not successfully matched.





\begin{figure*}[ht]
\centering
  \includegraphics[width=1\textwidth]{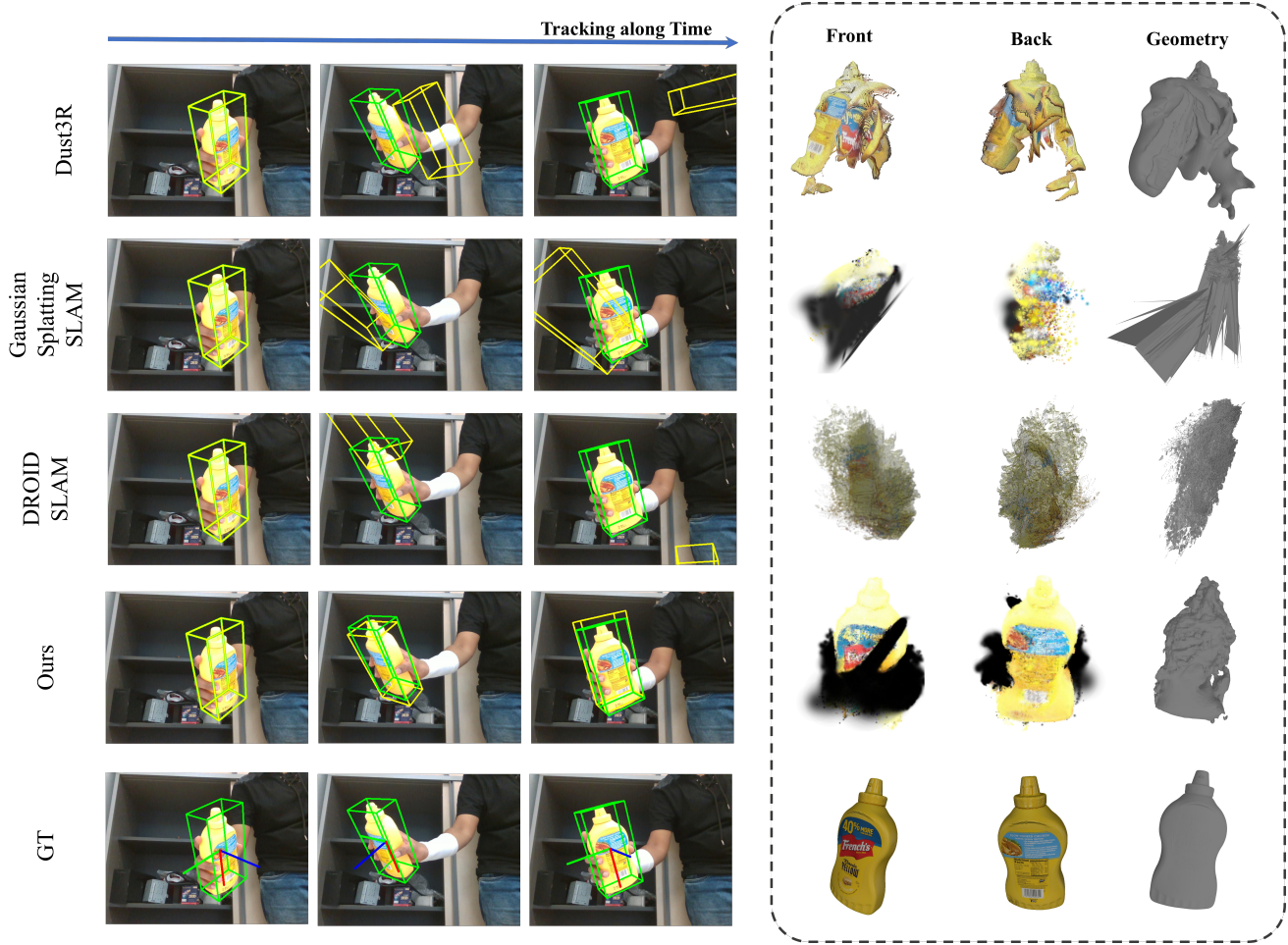}
  \caption{\textbf{Qualitative Comparison of GSGTrack and Baseline on HO3D.} Left: 6-DOF pose tracking with {\color{LimeGreen}green} and {\color{Yellow}yellow} boxes showing ground truth and estimated poses, respectively. Right: front and back views of reconstruction results, highlighting the object's geometric structure. Due to hand occlusions, black hand-shaped artifacts appear, obscuring parts of the object. Our reconstruction corrects the color divergence between ground truth and actual object colors seen in the video.}
  \label{fig:Ho3D_res}
\vspace{-3ex}
\end{figure*} 



\noindent\textbf{Confidence-aware Geometric Optimization.} In geometric graph pose optimization, we optimize only the current frame pose \( T_v \) to maintain consistency between the 3DGS and the graph poses, while historical keyframe poses are refined by the optimized Gaussian model. We simultaneously optimize the depth maps \( D_i \{ i \in [0,t]\} \) of all nodes and the edge transformation matrices \( T_{e2w} \) to align image pair points with the world coordinate system.

Our objective is to minimize the Chamfer Distance between graph nodes and geometry predicted by the generalized stereo matching network, thereby forming a consistent 3D representation. To reduce errors from inaccurate depth estimates, we incorporate the predicted confidence map into the loss function, weighting points by confidence. Thus, the geometry loss can be expressed as:

\begin{small}
\vspace{-2ex}
\setlength\abovedisplayskip{0.cm}
    \begin{equation}
        \mathcal{L}_{pg}=\sum_{e \in \mathcal{H}} \sum_{v \in \mathcal{E}_{e}} \sum_{i=1}^{h w} C_{i}^{v, e}\left\|\chi_{i}^{v}- T_{e2w} X_{i}^{v, e}\right\|,
    \end{equation}
\end{small}
where $\mathcal{H}$ denotes the structural graph, $e$ represents the graph edges, $\mathcal{E}_{e}$ refers to the two nodes connected by a directed edge, $C$ is the confidence map, $\chi_{i}^{v}$ is the pointmap of the node $v$, $T_{e2w}$ is the transformation matrix from the edge coordinate system to the world coordinate system, and $ X_{i}^{v, e}$ represents the pointmap of node $v$ in edge $e$. 


We employ the Gauss-Newton algorithm to optimize the structural graph, obtaining dense 2D-3D correspondences and the optimized object pose for the current frame. Geometric optimization is also performed on historical frames to correct potential errors in previous estimates.

\section{Experiments}
\label{sec:experiments}

\subsection{Experimental Setup}
\noindent \textbf{Datasets.} To evaluate our approach, we conduct extensive experiments on publicly available real-world datasets, OnePose~\cite{lin2022onepose} and HO3D~\cite{hampali2020honnotate}, which include multiple dynamic, object-centered videos. The OnePose dataset provides $20$FPS RGB videos of static objects at a resolution of $1920\times1440$. We utilize the SAM model to obtain object masks in the first frame and extracted scale information based on the provided 3D bounding box annotations. The HO3D dataset, on the other hand, contains RGBD videos centered on objects interacting with hands; we use only the RGB data for pose tracking and online reconstruction. We apply the annotated mask from BundleSDF bounding boxes and derive object scale estimates from ground-truth point clouds in the first frame.


\definecolor{best_result}{rgb}{1., 0.85, 0.70}
\definecolor{second_result}{rgb}{0.98, 0.78, 0.57}
\definecolor{third_result}{rgb}{1.0, 1.0, 0.56}

\begin{table*}[th]
\centering
\renewcommand{\arraystretch}{1.25}


\footnotesize
\caption{\textbf{Quantitative comparison on HO3D dataset.} We compared our method with baseline methods to evaluate the algorithm's capabilities in reconstruction and tracking.}
\label{exp:HO3D_Res_Table}
\setlength\tabcolsep{9.5pt}
\begin{tabular}{c|ccccc|ccccc}
\toprule 
\multirow[c]{2}{*}{ Method } & \multicolumn{5}{c|}{ ADD-S(\%)[0-0.3]m $\uparrow$ } & \multicolumn{5}{c}{ ADD(\%)[0-0.3]m $\uparrow$ } \\
 &  AP & MPM & SB & SM & Avg &  AP & MPM & SB & SM & Avg\\
 \midrule
Dust3R~\cite{wang2024dust3r} & 17.32 & 24.97 & 24.25 & 32.12 & 24.67 & 9.64 & 15.51 & 16.51 & 19.76 & 15.36 \\
Gaussian Splatting SLAM~\cite{matsuki2024gaussian} & 20.61 & 21.64 & 13.15 & 28.17 & 20.89 & 10.65 & 11.32 & 9.71 & 15.26 & 11.73 \\
DROID-SLAM~\cite{teed2021droid} & 3.21 & 0.32 & 5.34 & 9.67 & 4.64 & 0.77 &  0.17 & 3.55 & 5.64 & 2.53 \\
\textbf{GSGTrack} & 
\cellcolor{best_result}\textbf{70.04} & \cellcolor{best_result}\textbf{62.16} & \cellcolor{best_result}\textbf{63.70} & \cellcolor{best_result}\textbf{62.51} & \cellcolor{best_result}\textbf{64.60} & \cellcolor{best_result}\textbf{54.16} & \cellcolor{best_result}\textbf{43.81} & \cellcolor{best_result}\textbf{50.80} & 
\cellcolor{best_result}\textbf{51.83} & \cellcolor{best_result}\textbf{50.15} \\
\bottomrule
\end{tabular}

\vspace{0.1cm}


\setlength\tabcolsep{3.75pt}
\begin{tabular}{c|ccccc|ccccc|ccccc}
\toprule 
\multirow[c]{2}{*}{ Method } & \multicolumn{5}{c|}{ PSNR $\uparrow$ } & \multicolumn{5}{c|}{ SSIM $\uparrow$ } & \multicolumn{5}{c}{ Reconstruction CD~(cm) $\downarrow$} \\
 &  AP & MPM & SB & SM & Avg &  AP & MPM & SB & SM & Avg &  AP & MPM & SB & SM & Avg\\
 \midrule
Dust3R~\cite{wang2024dust3r} & — & — & — & — & — & — & — & — & — & — & 77.13 & 52.12 & 67.19 & 43.23 & 59.92 \\
Gaussian Splatting SLAM~\cite{matsuki2024gaussian} & 18.57 & 20.13 & 17.89 & 20.50 & 19.27 & 0.79 & 0.82 & 0.77 & 0.82 & 0.80 & 85.14 & 69.49 & 80.23 & 60.40 & 73.82 \\
DROID-SLAM~\cite{teed2021droid} & — & — & — & — & — & — & — & — & — & — & 150.33 & 130.80 & 81.86 & 100.87 & 115.97 \\
\textbf{GSGTrack} & 
\cellcolor{best_result}\textbf{26.70} & \cellcolor{best_result}\textbf{24.83} & \cellcolor{best_result}\textbf{25.20} & \cellcolor{best_result}\textbf{27.04} & \cellcolor{best_result}\textbf{25.92} & \cellcolor{best_result}\textbf{0.97} & \cellcolor{best_result}\textbf{0.96} & \cellcolor{best_result}\textbf{0.95} & 
\cellcolor{best_result}\textbf{0.97} & \cellcolor{best_result}\textbf{0.97} &
\cellcolor{best_result}\textbf{23.93} & \cellcolor{best_result}\textbf{15.72} & \cellcolor{best_result}\textbf{21.39} & \cellcolor{best_result}\textbf{19.20} & \cellcolor{best_result}\textbf{20.06} \\
\bottomrule
\end{tabular}
\vspace{-3ex}

\end{table*}

\noindent \textbf{Baselines.} To comprehensively compare GSGTrack, we include various types of benchmarks. To better evaluate the pose tracking capabilities of algorithms on RGB images, we compare several algorithms, including Droid-SLAM~\cite{teed2021droid}, a deep learning-based approach that jointly optimizes camera poses and scene structure; Gaussian Splatting SLAM (GS-SLAM) algorithm~\cite{matsuki2024gaussian}, which serves as our primary comparative baseline. GS-SLAM is implemented based on 3DGS to achieve simultaneous localization and mapping. We utilize depth information generated through a generalized depth matching model~\cite{wang2024dust3r} as input to assist Gaussian insertion in GS-SLAM. Droid-SLAM, on the other hand, inherently leverages optical flow and depth priors to assist sparse point cloud construction without requiring our depth prior information. Both algorithms are implemented using official open-source code.

To further compare reconstruction performance, we also evaluate against the Dust3R algorithm ~\cite{wang2024dust3r}, which performs end-to-end 3D reconstruction based on generalized stereo matching priors. For fair comparison, we restrict Dust3R’s image pairs to only use historical frames pointing to the current frame and prevent optimization of the historical frame's pose. Additionally, we compare the 3D reconstruction capabilities by using SfM~\cite{wu2011visualsfm} and 3DGS~\cite{kerbl2023gaussian} algorithms on the static scene dataset, OnePose.

\noindent \textbf{Metrics.} We separately evaluate the quality of pose tracking and reconstruction accuracy. For 6-DOF object pose, we calculate the area under the ADD and ADD-S metrics curves (0 to 0.3m) using the actual object geometry~\cite{he2022fs6d,Xiang2018PoseCNN}. Given that the OnePose dataset lacks ground-truth object geometries, we indirectly measure the pose tracking accuracy for static objects by evaluating the precision of the camera trajectory~\cite{sturm2012evaluation}. For 3D reconstruction, PSNR and SSIM metrics are used to assess the quality of appearance~\cite{wang2004image}. Additionally, to evaluate the accuracy of object shape reconstruction, we compute the Chamfer Distance between the final reconstructed mesh and the ground-truth mesh defined in the reference coordinate system of the first video frame~\cite{Wen2023BundleSDF}.


\subsection{Implementation Details}

For each video frame, we use object segmentation for scaling and cropping to focus on the object. Following 3DGS~\cite{kerbl2023gaussian}, both time-critical rasterization and gradient computation are implemented using CUDA. We implement the graph-based geometric optimization using PyTorch, with object-specific structural optimization carried out using the Adam optimizer. Coarse optimization runs for 300 iterations, followed by 125 pose refinement iterations per frame after pose estimation. All experiments use an NVIDIA GeForce RTX 3090. To ensure generalizability, 
We use the officially released network weights of the Dust3R algorithm, which are not trained on HO3D or OnePose datasets.

\subsection{Results on the HO3D Dataset}



The quantitative results of the comparison on the HO3D dataset are shown in ~\cref{exp:HO3D_Res_Table}. The proposed method demonstrates significant improvements in 6DoF object pose tracking and 3D reconstruction. For the DROID-SLAM algorithm, working in object-centered scenes reduces the availability of textures and geometric cues for tracking in the images. This environment significantly diminishes the reliability of the optical flow and depth priors on which the algorithm depends, resulting in an overall decline in performance. Both the Dust3R and GS-SLAM are enhanced with generalized stereo matching to adopt stronger 3D geometric priors. However, these algorithms build a globally optimized model that does not adequately handle noise and errors in the depth priors. Consequently, as the inference progresses over multiple frames, errors quickly accumulate within the global model, causing continual degradation in performance and eventually leading to tracking failure.

\cref{fig:Ho3D_res} presents qualitative comparisons with other methods. Despite various challenges (\eg, severe hand occlusions, self-occlusions, frames lacking texture and geometric cues, and strong light reflections), our algorithm successfully tracks the object's 6DoF pose and achieves a significantly high-quality 3D appearance representation. Notably, the appearance of the reconstructed 3D object in our approach better aligns with the texture and color information of the source object in the scene, compared to the ground truth reconstructed from scans.

\subsection{Results on the OnePose Dataset}
\begin{table}[t]
\centering
\renewcommand{\arraystretch}{1.15}
\footnotesize
\caption{Quantitative comparison on the OnePose dataset.}
\vspace{-1.5ex}
\label{exp：OnePose_Res}
\setlength\tabcolsep{1pt}
\begin{tabular}{c|cccc}
\toprule
\multirow{1}{*}{Method}  & APE (cm)$\downarrow$ & RPE (cm)$\downarrow$ & PSNR$\uparrow$ & SSIM$\uparrow$ \\
\midrule
Dust3R~\cite{wang2024dust3r} & 30.42 & 25.29 & \textemdash & \textemdash \\
Gaussian Splatting SLAM~\cite{matsuki2024gaussian} & 10.28 & 9.62 & 19.27 & 0.85 \\
DROID-SLAM~\cite{teed2021droid} & 8.57 & \cellcolor{best_result}\textbf{6.94} & \textemdash & \textemdash \\
SfM~\cite{wu2011visualsfm}+3DGS~\cite{kerbl2023gaussian} & \textemdash & \textemdash & 21.43 & 0.87 \\
\textbf{GSGTrack} & \cellcolor{best_result}\textbf{7.36} & \textbf{8.79} & \cellcolor{best_result}\textbf{23.22} & \cellcolor{best_result}\textbf{0.90}\\
\bottomrule
\end{tabular}
\end{table}

The quantitative results on the OnePose dataset are shown in \cref{exp：OnePose_Res}. This dataset consists of object-centered static scenes, where we use a video segmentation network to isolate the object. Our algorithm exhibits superior global tracking compared to previous SLAM algorithms, especially when background information is excluded. However, tracking stability at finer details fluctuates, yielding weaker RPE metrics compared to DROID-SLAM.
We also compare with the SfM+3DGS method, which reconstructs poses from full-scene, unsegmented views using 3DGS. Due to scene complexity, poses derived from SfM frequently show deviations, causing misalignment between reconstructed and ground-truth views during object reconstruction and reducing overall quality. This further underscores GSGTrack’s advantages in object-centered reconstruction.

\subsection{Ablation Study}

\begin{table}[t!]
\centering
\renewcommand{\arraystretch}{1.15}
\footnotesize
\caption{Ablation study of different settings of our methods.}
\vspace{-1.5ex}
\label{exp:Ablation_Res}
\setlength\tabcolsep{4pt}
\begin{tabular}{c|cccc}
\toprule
\multirow{1}{*}{Method}  & \makecell{ADD-S\%\\$\le$0.3m$\uparrow$} & \makecell{ADD\%\\ $\le$0.3m$\uparrow$} & PSNR$\uparrow$ & SSIM$\uparrow$ \\
\midrule
\emph{w/o} Tracking & 32.22 & 23.14 & 14.54 & 0.88 \\
\emph{w/o} Silhouette Loss & 56.31 & 42.16 & 24.21 &  \cellcolor{best_result}\textbf{0.97} \\
\emph{w/o} Geometric Graph & 25.20 & 15.93 & 22.77 & 0.95 \\
\emph{w/o} Image Pruning & 50.99 & 39.44 & 23.59 & \cellcolor{best_result}\textbf{0.97} \\
\emph{w/o} Geometric Optimization & 51.08 & 32.96 & 24.30 & 0.96 \\
\textbf{Ours} & \cellcolor{best_result}\textbf{62.51} & \cellcolor{best_result}\textbf{51.83} & \cellcolor{best_result}\textbf{27.04} & \cellcolor{best_result}\textbf{0.97}\\
\bottomrule
\end{tabular}
\vspace{-3ex}
\end{table}



\begin{figure}
    \centering
    \includegraphics[width=\linewidth]{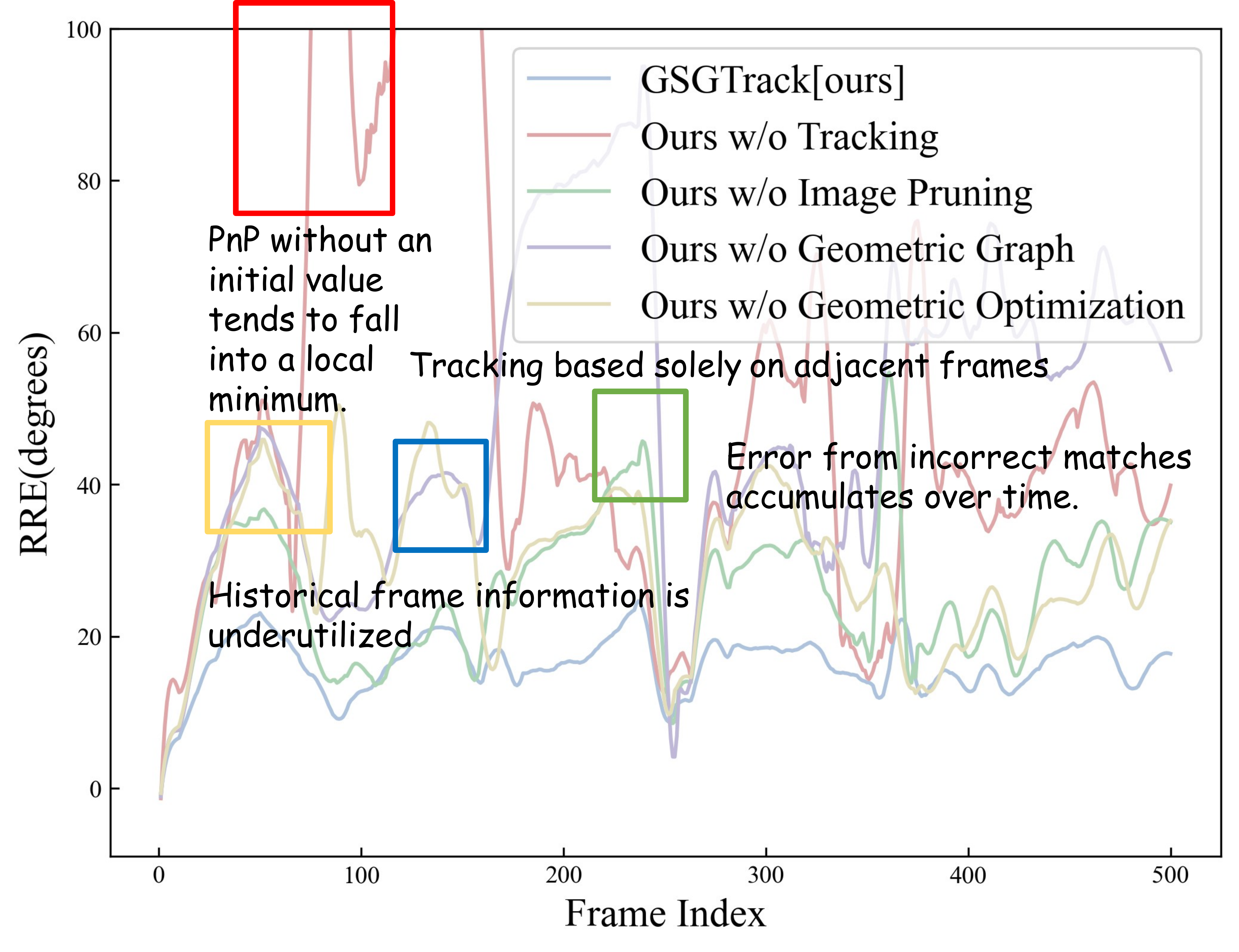}
    \vspace{-4ex}
    \caption{We visualize the Relative Rotation Error (RRE) of different settings of our method.}
    \label{fig:RRE_Curve}
    \vspace{-2ex}
\end{figure}

We conduct extensive ablation studies on the HO3D dataset to validate the effectiveness of our proposed strategies, with results presented in \cref{exp:Ablation_Res} and \cref{fig:RRE_Curve}.

\noindent\textbf{Pose Tracking.} As mentioned previously, we track poses across frames to initialize the object pose. To assess the impact of this strategy, we remove pose tracking during initialization (\emph{w/o} Tracking) and instead use a PnP algorithm to estimate poses. The results demonstrate that tracking quickly drifts due to inaccurate initialization.

\noindent\textbf{Silhouette Loss.} We introduce a differentiable silhouette loss to mitigate errors caused by the simple photometric loss. To evaluate its effectiveness, we exclude this component from the experiments (\emph{w/o} Silhouette Loss). The results show a clear performance decline, with ADD-S@0.3d and ADD@0.3d decreasing by 10\% and 19\% respectively.

\noindent\textbf{Graph-based Geometric Optimization.} To validate the importance of graph-based optimization, we remove it and rely only on the latest frame for updates (\emph{w/o} Geometric Graph), resulting in error accumulation over time. We also examine the impact of removing the image pair pruning strategy (\emph{w/o} Image Pruning), which leads to degraded performance due to failed edges. Finally, omitting gradient descent optimization on the geometric graph shows that geometric optimization is essential for improving performance.

\begin{figure}
    \centering
    \includegraphics[width=\linewidth]{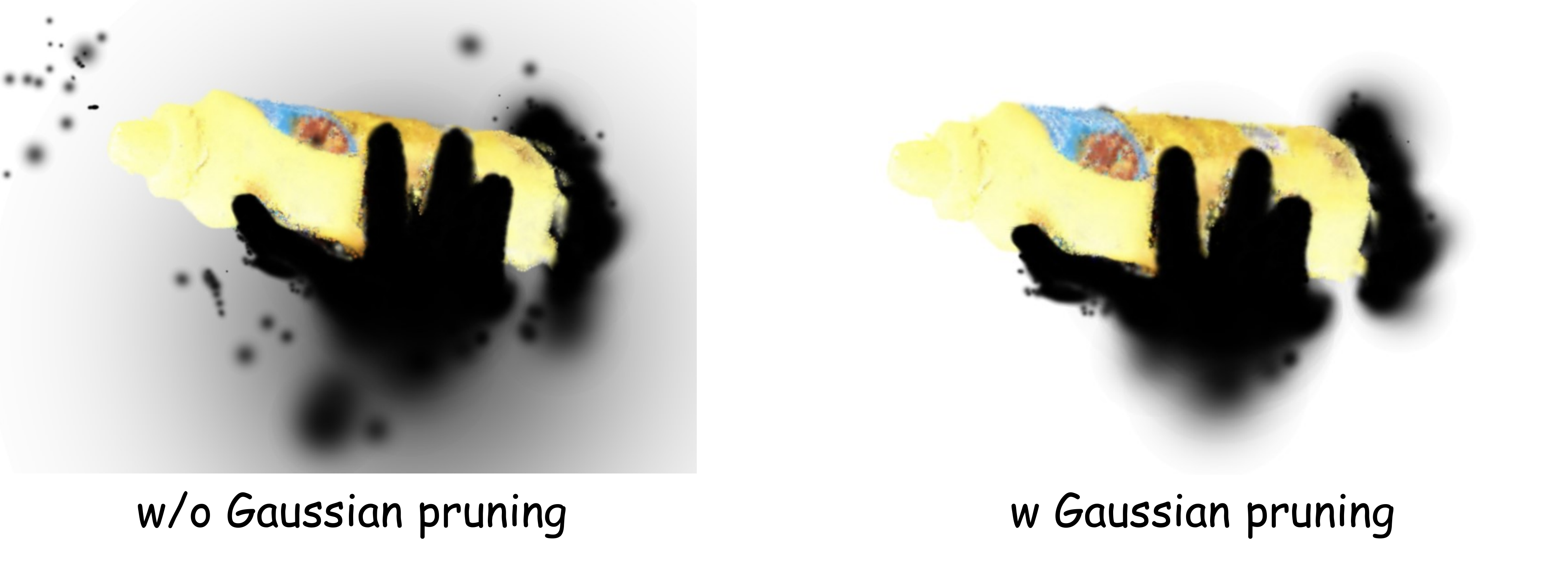}
    \vspace{-4ex}
    \caption{Impact of our Gaussian pruning strategy on reconstruction quality. Our strategy significantly enhances geometric accuracy and effectively eliminates floaters.}
    \label{fig:prune}
    \vspace{-4ex}
\end{figure}

\noindent\textbf{Gaussian Pruning.} To address geometric inaccuracies, we propose a Gaussian pruning strategy. To validate its effect, we remove it from the pipeline and provide qualitative comparisons in \cref{fig:prune}. The results reveal that our strategy improves geometric accuracy and effectively prunes floaters.

\section{Conclusion}

This paper addresses a challenging problem: 6D pose tracking of unknown objects from RGB videos without accurate depth information. 
To address this, we introduce GSGTrack, a novel method that leverages Gaussian splatting to enhance pose tracking. Our approach employs a joint optimization framework to simultaneously refine object poses and their 3D representation. To manage continuously incoming video frames during tracking, we develop an online 3DGS representation, enabling incremental object reconstruction. Furthermore, we propose a graph-based geometric optimization framework that integrates an image pair pruning strategy and a confidence-aware optimization strategy to improve accuracy and robustness. Extensive experiments across multiple datasets show that our framework achieves robust pose tracking and accurate 3D reconstruction, even with inaccurate initial geometric information.

{
    \small
    \bibliographystyle{ieeenat_fullname}
    \bibliography{main}
}

\clearpage
\setcounter{page}{1}
\maketitlesupplementary

In this supplementary material, we provide the implementation details of the experiments, along with the algorithm evaluation metrics and comprehensive information about the datasets. Furthermore, we provide qualitative results for challenging scenarios in the datasets, analyze the limitations of the proposed algorithm, and discuss its broader impact.

\section{Implementation details}
\label{sec:implementation}

\subsection{Data Preprocessing}

During the data preprocessing stage, for the segmented video image \( F_t \), we first enlarge the image to reduce the relative proportion of background noise. Subsequently, using the segmented mask as a reference, we crop the image so that the projection center of the object in the 2D image aligns as closely as possible with the center of the image, thereby reducing the difficulty of geometric estimation. During the image processing, we simultaneously adjust the camera's intrinsic parameters to maintain the validity of the solved camera extrinsic parameters. Specifically, the transformation matrix for the camera's intrinsic parameters \( M \) is as follows:

\begin{small}
    \begin{equation}
        M=\left[\begin{array}{ccc}
K & 0 & -K x_{0}+\frac{w^{\prime}}{2} \\
0 & K & -K y_{0}+\frac{h^{\prime}}{2} \\
0 & 0 & 1
\end{array}\right],
    \end{equation}
\end{small}
where $(h^{\prime}, w^{\prime})$ represents the dimensions of the target image, $K$ is the scaling factor, and $(x_0, y_0)$ denotes the center of the object in the 2D image.

\subsection{System Details and Hyperparameters}


During the pose initialization process (Sec 3.4 in the main manuscript), if the pose of the previous frame is not available as a direct reference(\eg missing detection by the segmentation or object reappearing after complete occlusion), the algorithm first performs generalized stereo matching between the current frame and historical keyframes in the pose graph to estimate the pose of the current frame. This estimated pose is then used as the initial pose for the tracking process. If the tracking state of the previous frame is valid, the initial pose of the current frame's tracking process is directly set to the pose of the previous frame. In experiments, 100 iterations of tracking are executed, and if the magnitude of the pose update falls below $10^{-4}$, the iterations are terminated early.


During the geometric graph optimization process, we limit the optimization to two layers: optimizing the pose and depth map of the current frame simultaneously, as well as the depth maps of the reference frames. Pixels with a confidence score lower than 2 are excluded from the computation of $L_{pg}$. The geometric graph is optimized for approximately 300 iterations in total.


During the training process of the online 3D Gaussian Splatting algorithm, considering that the geometry optimization algorithm has already provided good initial values for the 3D Gaussian points, the learning rate for the Gaussian point positions is reduced to 0.000032. In object-level scenes, the Gaussian point size threshold is set to 3. For the initial frame view of each object, the algorithm trains for 325 iterations to initialize the 3D Gaussian Splatting algorithm. Subsequently, Gaussian optimization is conducted for 125 iterations per video frame, incrementally reconstructing the 3D Gaussian representation of the object online while optimizing the object's 6-DoF pose. Every 25 iterations, the algorithm executes a density control strategy consistent with the classical 3D Gaussian Splatting algorithm. During the final iteration of Gaussian optimization, the algorithm applies a Gaussian pruning strategy, as described in (Sec 3.3), to remove geometrically inaccurate 3D Gaussian points.

\subsection{Keyframing}

To ensure the efficiency of algorithm execution, it is impractical to perform the same optimization process for every frame. Therefore, in our implementation, we calculate the rotational geodesic distance of each frame relative to the nodes in the geometric pose graph. This approach ensures that the keyframes added to the geometric graph provide novel information, including texture details, viewpoint diversity, and scale variations of the object. Non-keyframes, on the other hand, are only initialized with a pose estimate (Sec 3.4) and do not participate in subsequent geometric graph optimization or Gaussian model refinement processes.

\subsection{Visualization}

For the reconstructed 3D Gaussian Splatting model, we visualize it using supersplat~\cite{SuperSplat}. For the 3D pointmaps, we select points with confidence greater than 2, merge them in the world coordinate system to form a unified object point cloud, and apply voxel-based uniform downsampling to obtain a consistent point cloud representation of the object. The mesh model used for visualization is generated from the object point cloud by reconstructing the object surface using the Poisson reconstruction algorithm.

\subsection{Gradient Derivation}

For efficiency, 3DGS~\cite{kerbl2023gaussian} employs CUDA-based rasterization, requiring explicit computation of parameter derivatives. Consequently, the chain rule is applied to differentiate Eq.(19), yielding partial derivatives as follows:
\begin{align}
\frac{\partial \boldsymbol{\mu}^{\prime}}{\partial \boldsymbol{T}} & =\frac{\partial \boldsymbol{\mu}^{\prime}}{\partial \boldsymbol{\mu}} \frac{\mathcal{D} \boldsymbol{\mu}}{\mathcal{D} \boldsymbol{T}}, \\
\frac{\partial \boldsymbol{\Sigma}^{\prime}}{\partial \boldsymbol{T}} & =\frac{\partial \boldsymbol{\Sigma}^{\prime}}{\partial \mathbf{J}} \frac{\partial \mathbf{J}}{\partial \boldsymbol{\mu}} \frac{\mathcal{D} \boldsymbol{\mu}}{\mathcal{D} \boldsymbol{T}}+\frac{\partial \boldsymbol{\Sigma}^{\prime}}{\partial \mathbf{W}} \frac{\mathcal{D} \mathbf{W}}{\mathcal{D} \boldsymbol{T}}.
\end{align}

Following gaussian splatting SLAM~\cite{matsuki2024gaussian}, we derived the minimal Jacobian matrix on the manifold using Lie algebra and explicitly computed the derivatives of the camera pose.
\begin{small}
\begin{equation}
    \frac{\mathcal{D} \boldsymbol{\mu}}{\mathcal{D} \boldsymbol{T}}=\left[\begin{array}{ll}\boldsymbol{I} & -\boldsymbol{\mu}^{\times}\end{array}\right], \frac{\mathcal{D} \mathbf{W}}{\mathcal{D} \boldsymbol{T}}=\left[\begin{array}{ll}\mathbf{0} & -\mathbf{W}_{i, 1}^{\times} \\\mathbf{0} & -\mathbf{W}_{i, 2}^{\times} \\\mathbf{0} & -\mathbf{W}_{i, 3}^{\times}\end{array}\right].
\end{equation}
\end{small}

\section{Metrics}

To evaluate the results of the algorithm, we assess both 6DoF pose tracking and object reconstruction. For the 6DoF pose tracking results, we compute the Area Under the Curve (AUC) percentages for the ADD and ADD-S metrics,

\begin{small}
\setlength\abovedisplayskip{0.cm}
    \begin{equation}
        \text { ADD }=\frac{1}{|\mathcal{M}|} \sum_{x \in \mathcal{M}}\|(R x+t)-(\tilde{R} x+\tilde{t})\|_{2},
    \end{equation}
\end{small}
\begin{small}
\setlength\abovedisplayskip{0.cm}
    \begin{equation}
        \text { ADD-S } =\frac{1}{|\mathcal{M}|} \sum_{x_{1} \in \mathcal{M}} \min _{x_{2} \in \mathcal{M}}\left\|\left(R x_{1}+t\right)-\left(\tilde{R} x_{2}+\tilde{t}\right)\right\|_{2},
    \end{equation}
\end{small}
where \(\mathcal{M}\) is the object model. Since the CAD model of the novel, unknown object is unavailable for defining a coordinate system, we utilize the ground-truth pose from the first frame to establish the canonical coordinate frame for each video, enabling pose evaluation.

For 3D reconstruction evaluation, the object's 3D model is projected onto 2D images and rendered from corresponding viewpoints. The projections are then compared with ground-truth images using PSNR and SSIM metrics.
To evaluate the accuracy of 3D reconstruction shapes, we followed BundleSDF~\cite{Wen2023BundleSDF}. Specifically, we assessed the Chamfer Distance between the final geometrically optimized point cloud and the downsampled point cloud from the ground-truth mesh. The symmetric formula used is as follows:
\begin{small}
\setlength\abovedisplayskip{0.cm}
    \begin{equation}
        \begin{aligned}
        d_{C D}= & \frac{1}{2\left|\mathcal{M}_{1}\right|} \sum_{x_{1} \in \mathcal{M}_{1}} \min _{x_{2} \in \mathcal{M}_{2}}\left\|x_{1}-x_{2}\right\|_{2}+ \\
        & \frac{1}{2\left|\mathcal{M}_{2}\right|} \sum_{x_{2} \in \mathcal{M}_{2}} \min _{x_{1} \in \mathcal{M}_{1}}\left\|x_{1}-x_{2}\right\|_{2}.
        \end{aligned}
    \end{equation}
\end{small}
In the shape evaluation process, we downsampled the point cloud to a uniform resolution of 5 mm. 

\section{Datasets}


\begin{table}[]
\footnotesize
\centering
\setlength\tabcolsep{3pt}
\caption{\textbf{Scene sequences of HO3D and OnePose}.} 
\label{table:data_seq}
\renewcommand{\arraystretch}{1.25}
\begin{tabular}{cccc}
\hline
\multicolumn{2}{c}{\textbf{HO3D}} & \multicolumn{2}{c}{\textbf{OnePose}} \\ \hline
\multirow{2}{*}{\textbf{Pitcher Base}} & AP11 & \textbf{0500-Chocfranzzi-Box} & Choc-01 \\
 & AP14 & \textbf{0501-Matchafranzzi-Box} & Mat-01 \\
 \multirow{1}{*}{\textbf{Potted Meat Can}} & MPM14 & \textbf{0518-Jasmine-Box} & Jas-01 \\
 \multirow{2}{*}{\textbf{Bleach Cleanser}} & SB11 & \textbf{0535-Odbmilk-Box} & Odb-01 \\
 & SB13 & \textbf{0543-Brownhouses-Others} & Brown-01 \\
 \multirow{1}{*}{\textbf{Mustard Bottle}} & SM1 &  &  \\
\hline
\end{tabular}
\vspace{-.1cm}
\end{table}

\begin{figure}
    \centering
    \includegraphics[width=\linewidth]{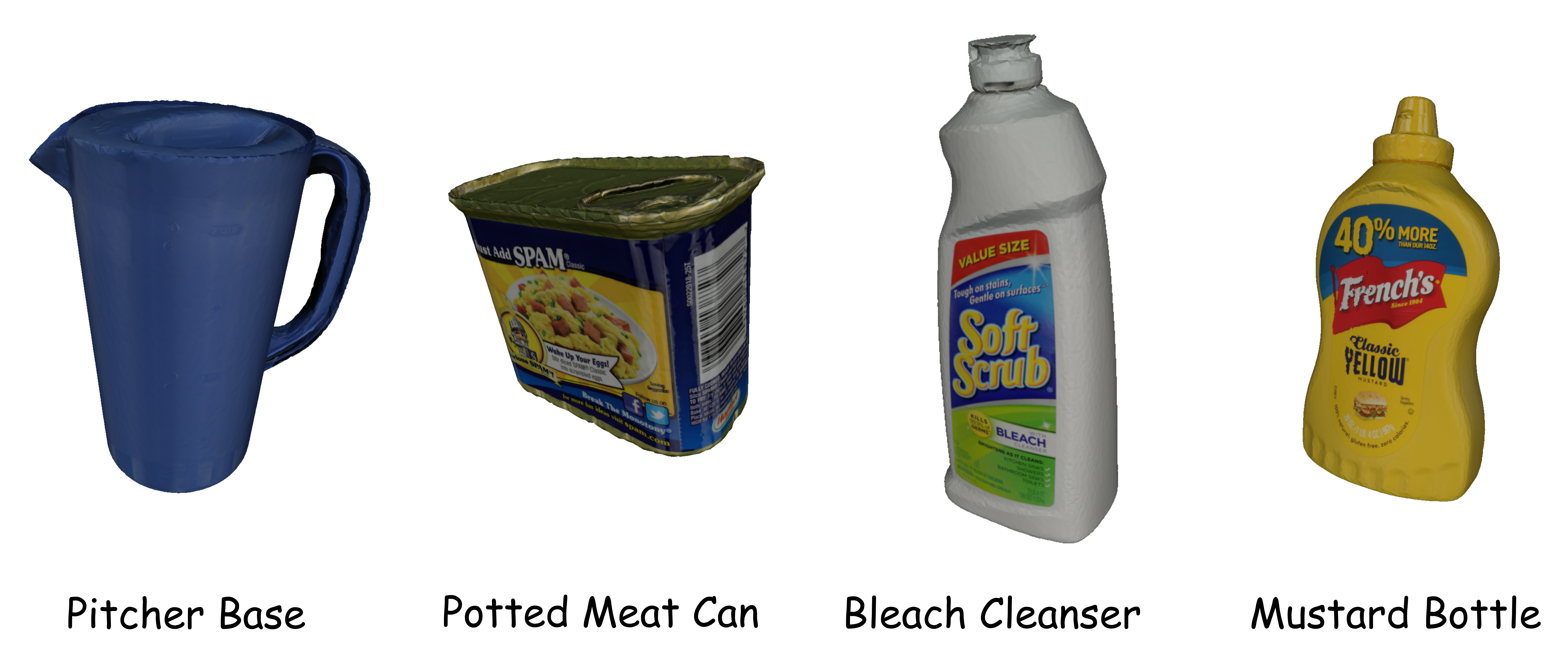}
    \vspace{-4ex}
    \caption{Visualization for the objects of in HO3D dataset}
    \label{fig:HO3D_Overview}
    \vspace{-2ex}
\end{figure}

\begin{figure}
    \centering
    \includegraphics[width=\linewidth]{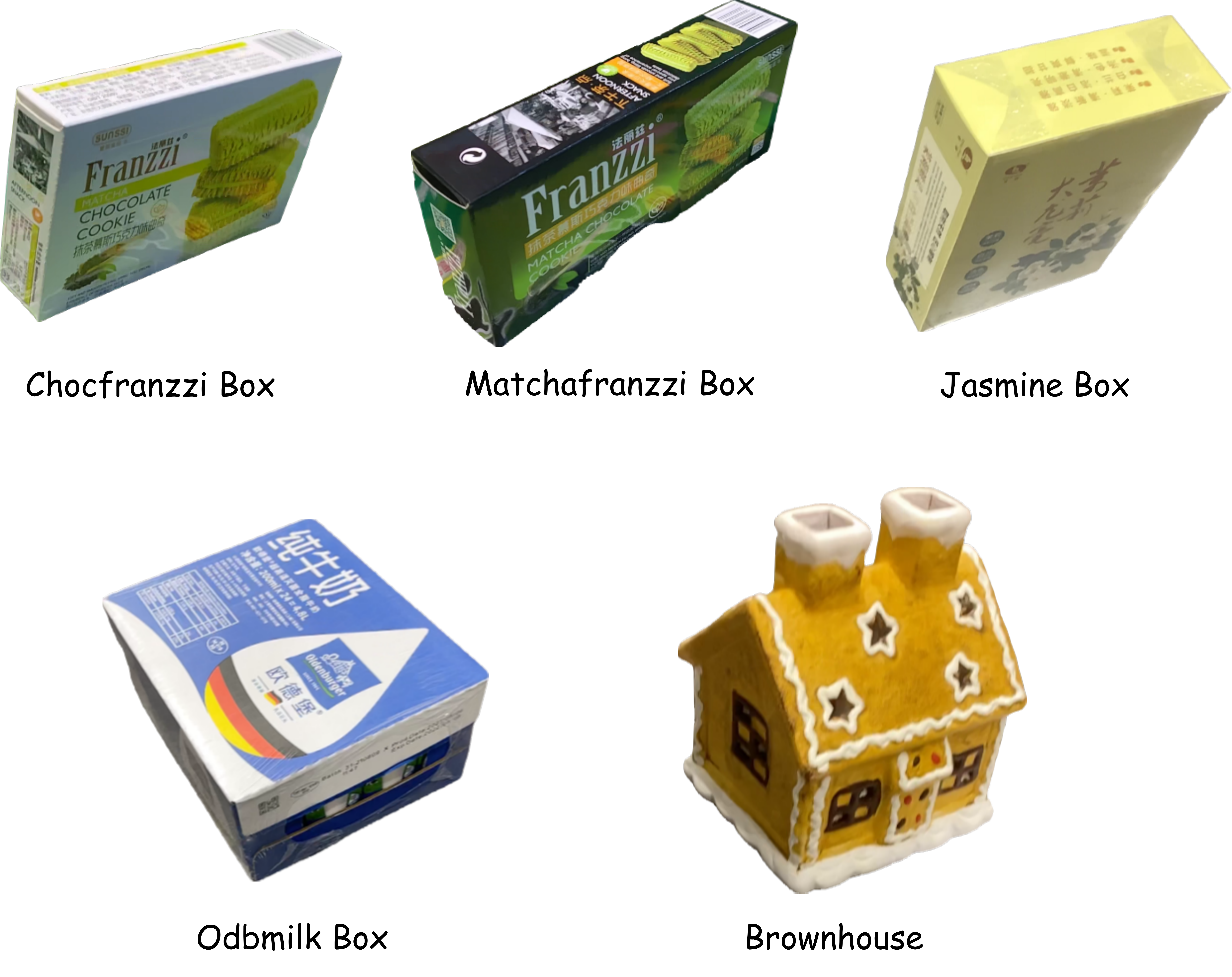}
    \vspace{-4ex}
    \caption{Visualization for the objects of in OnePose dataset}
    \label{fig:OnePose_Overview}
    \vspace{-2ex}
\end{figure}

\begin{figure*}[ht]
\centering
  \includegraphics[width=1\textwidth]{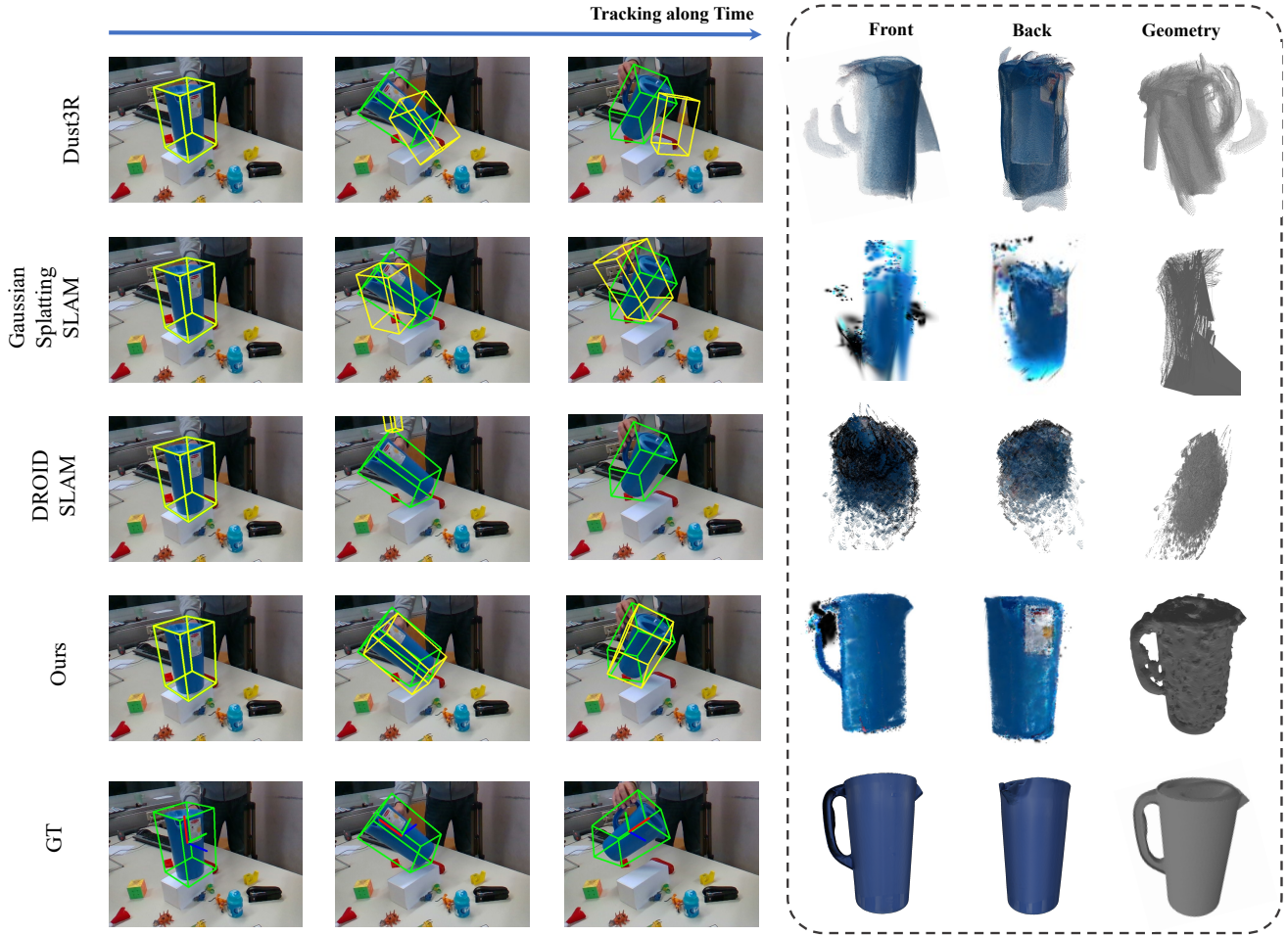}
  \caption{\textbf{Qualitative Comparison of GSGTrack and Baseline on HO3D(Seq-AP11).} Left: 6-DOF pose tracking with {\color{LimeGreen}green} and {\color{Yellow}yellow} boxes showing ground truth and estimated poses, respectively. Right: front and back views of reconstruction results, highlighting the object's geometric structure. A blue pitcher with low texture is presented. The qualitative results demonstrate that, compared to the baseline algorithm, our method exhibits significantly enhanced robustness in handling low-texture objects.}
  \label{fig:AP11_res}
\end{figure*}

\begin{figure*}[ht]
\centering
  \includegraphics[width=0.85\textwidth]{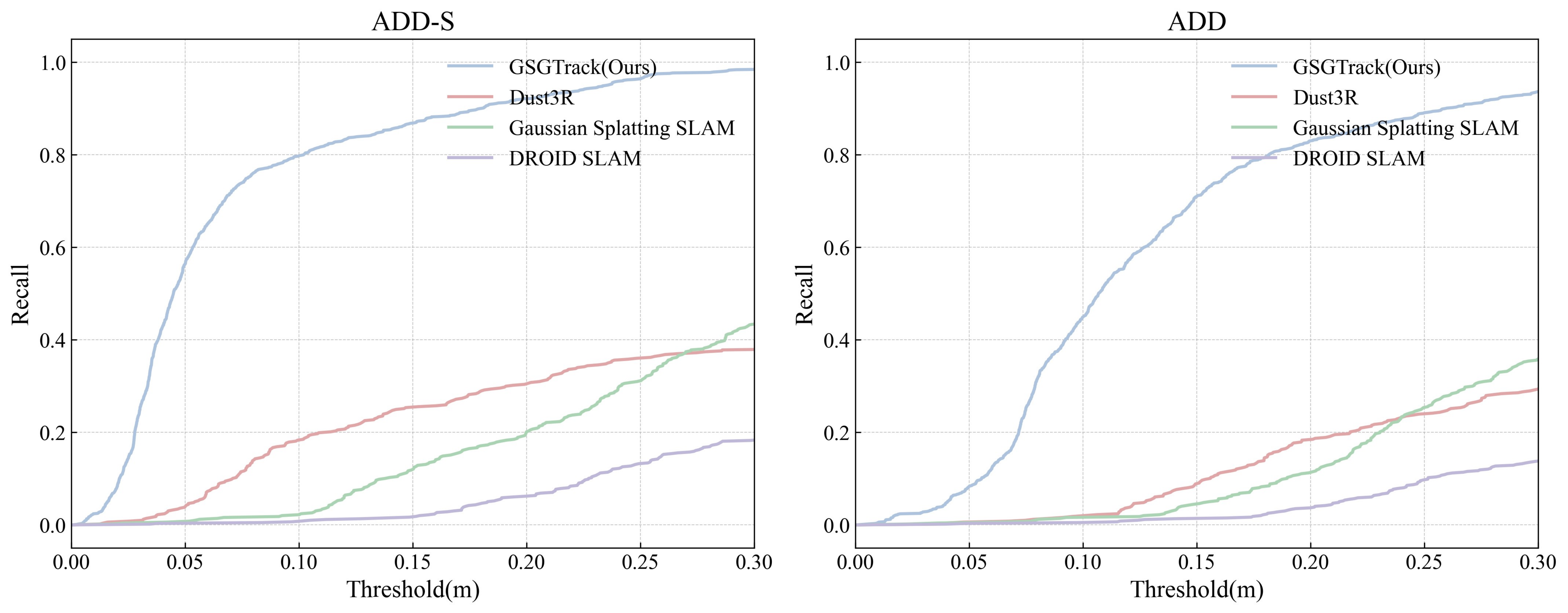}
  \caption{Recall curve of ADD-S (left) and ADD (right) metric on HO3D Dataset..}
  \label{fig:AUC}
\end{figure*}

\begin{figure*}[ht]
\centering
  \includegraphics[width=1\textwidth]{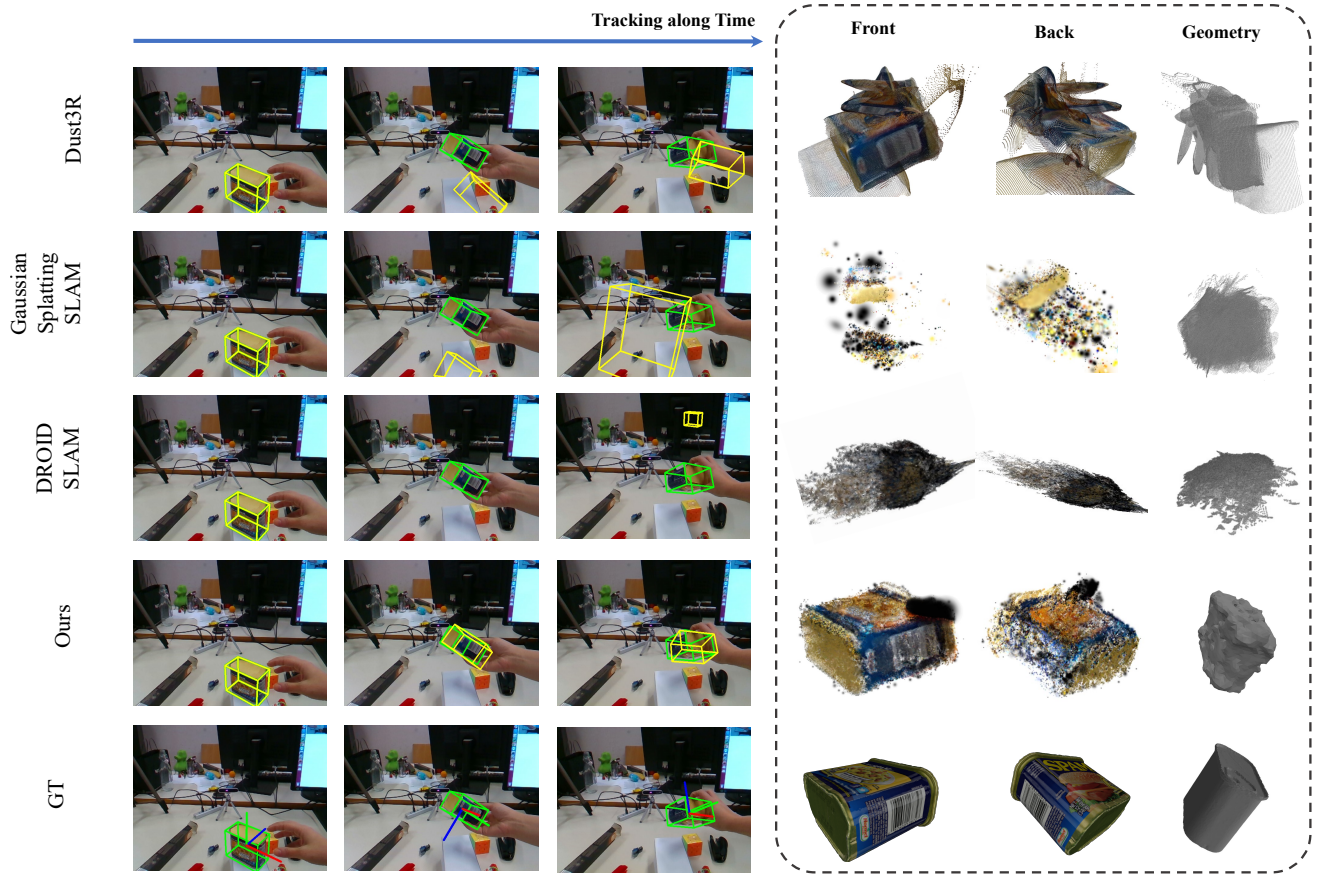}
  \caption{\textbf{Qualitative Comparison of GSGTrack and Baseline on HO3D(Seq-MPM14).} Left: 6-DOF pose tracking with {\color{LimeGreen}green} and {\color{Yellow}yellow} boxes showing ground truth and estimated poses, respectively. Right: front and back views of reconstruction results, highlighting the object's geometric structure. A Potted Meat Can object partially occluded by a hand is presented. The qualitative results demonstrate that, compared to the baseline algorithm, our approach exhibits significantly enhanced robustness in handling scenarios with occlusion.}
  \label{fig:MPM14_res}
\end{figure*}

\begin{figure*}[ht]
\centering
  \includegraphics[width=0.8\textwidth]{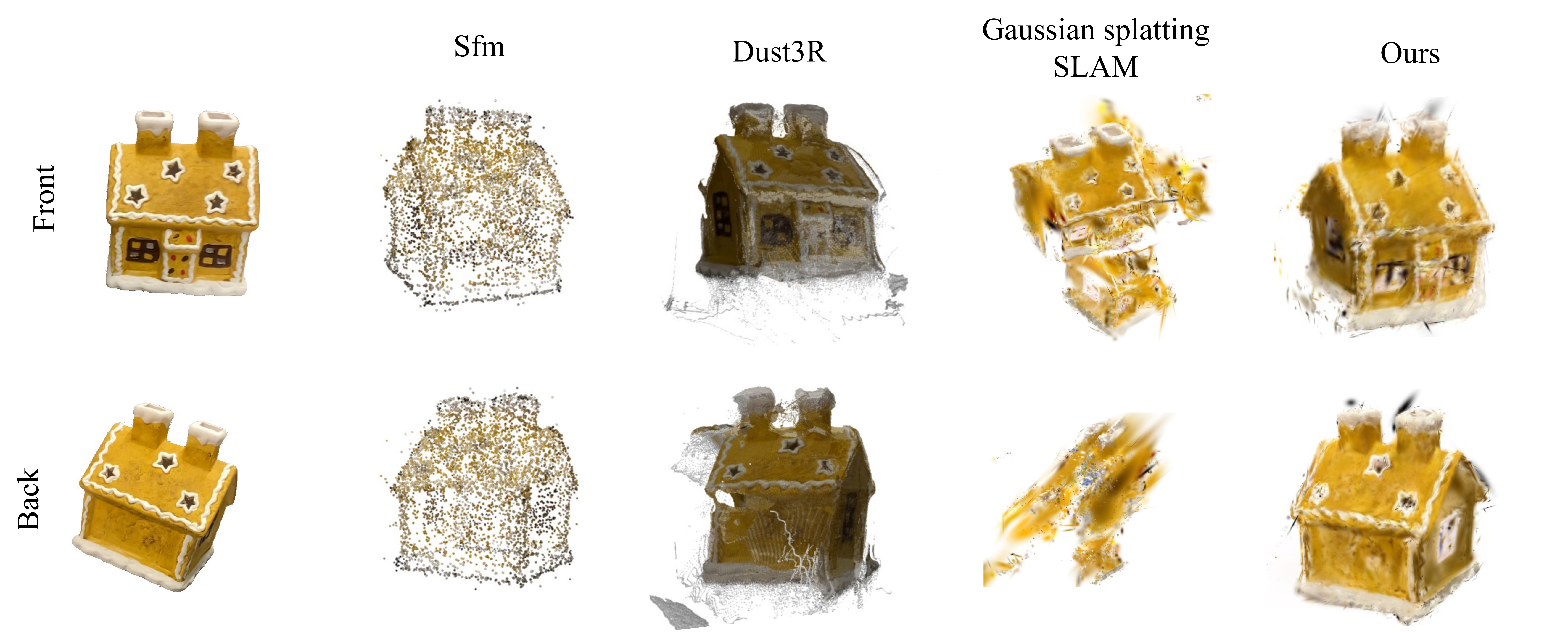}
  \caption{\textbf{Qualitative Comparison of GSGTrack and Baseline on OnePose.} Our method demonstrates superior object reconstruction quality.}
  \label{fig:OnePose_Res}
\end{figure*}

As shown in the \cref{fig:HO3D_Overview}, we selected 6 representative video sequences from the HO3D dataset, which include 4 dynamic objects. Each scene contains approximately 1,000 frames of data, featuring dynamic objects and hands interacting with them. The scale information for the first frame was calculated using the ground-truth depth values provided by the dataset. Based on this, we conducted experiments on the dataset. \cref{fig:AUC} presents the ADD-S and ADD recall curves obtained from these experiments, while \cref{fig:AP11_res} and \cref{fig:MPM14_res} show the qualitative results for this dataset. It can be observed that our method outperforms all other methods in both qualitative and quantitative metrics on the HO3D dataset. For the OnePose dataset, we selected 5 video sequences as illustrated in \cref{fig:OnePose_Overview}, containing five static objects. Each scene comprises approximately 500 frames, while \cref{fig:OnePose_Res} show the qualitative results for this dataset. . The significant variations across scenes serve as an accurate indicator of the reconstruction capabilities of current methods. The indexing of the scene sequences is provided in \cref{table:data_seq}.



\section{Limitation}

Although our method demonstrates greater robustness than the baseline algorithm in handling low-textured objects and occlusions (as shown in \cref{fig:AP11_res} and \cref{fig:MPM14_res}), it performs poorly when dealing with uniformly colored objects that lack geometric, color, or texture features. The method relies on the first frame of the video to initialize the object's local coordinate system, making it sensitive to the quality of initial matching. Such matching can be compromised by segmentation errors, lighting issues, or insufficient texture in the initial viewpoint. For instance, in the AP10 sequence of the HO3D dataset, the absence of geometric cues in the first frame significantly degrades performance. Additionally, the method assumes that each 2D image point corresponds to a 3D world point, limiting its applicability to transparent objects.

\section{Broader Impact}

The GSGTrack framework introduces a significant leap forward in the field of 6-DoF pose tracking and 3D object reconstruction, particularly for applications relying solely on monocular RGB video data. By eliminating the reliance on accurate depth information, the proposed approach offers broader accessibility and applicability in scenarios such as robotic manipulation, augmented reality, and autonomous systems, where lightweight and cost-effective sensing solutions are required. The novel 3D Gaussian Splatting representation and integrated graph-based geometric optimization framework enable robust pose tracking and high-fidelity object reconstruction, advancing theory and offering practical tools for interdisciplinary applications.

\end{document}